\documentclass[letterpaper, 10 pt, conference]{ieeeconf}  

\usepackage{graphicx}
\usepackage{amsmath}
\usepackage{enumerate}
\usepackage{acro}
\usepackage{subfigure}
\usepackage{caption}
\usepackage{lipsum}

\let\labelindent\relax
\usepackage{enumitem}                                               

\IEEEoverridecommandlockouts                             
\title{\LARGE \bf
Neural Networks Model for Travel Time Prediction Based on OD Travel Time Matrix
}

\author{Ayobami Ephraim Adewale$^{1}$, Amnir Hadachi$^{2}$
\thanks{keywords: O-D matrix, Deep learning, Neural Networks, Intelligent Transportation System, Public Transport}
}

\begin{document}

\maketitle
\thispagestyle{empty}
\pagestyle{empty}

\begin{abstract}

Public transportation system commuters are often interested in getting accurate travel time information to plan their daily activities. However, this information is often difficult to predict accurately due to the irregularities of road traffic, caused by factors such as weather conditions, road accidents, and traffic jams. In this study, two neural network models namely multi-layer(MLP) perceptron and long short-term model(LSTM) are developed for predicting link travel time of a busy route with input generated using Origin-Destination travel time matrix derived from a historical GPS dataset. The experiment result showed that both models can make near-accurate predictions however, LSTM is more susceptible to noise as time step increases.  

\end{abstract}

\section{INTRODUCTION}

In the last decade, public transportation system have been used to reduce traffic congestion, reduce the emission of greenhouse gas that are harmful to the environment, improve access to opportunities within connected cities, boost economy growth and lastly, improve quality of life in general. Despite it's benefits, most people are still reluctant to take public bus and prefer to use their private vehicles. This is quite understandable because the system is easily affected by weather, traffic signals, traffic fluctuations, peak hours and road incidents which often leads to delay and irregularities in schedules.  Based on this, there have been increasing demand for scientific techniques to solve the lingering problems.

In this paper, the main objective is to apply the concept of Origin Destination travel time matrix and Deep learning techniques in the prediction of link travel times in public transport. The second objective is to compare the prediction accuracy of multiple-layer perceptron (MLP) neural network (NN) with a time series neural network like long short-term memory (LSTM). Although different research involving the use of LSTM in travel time prediction already exist, very few study compare the prediction accuracy with that of MLP. Also, unlike in other studies, different independent variables were considered when creating the LSTM NN used in this study.  

The paper is organized as follows,  section 2 gives a review of state-of-the-art methods related to Neural Network methods, section 3 gives a structure of the two prediction models used in this paper. In section 4, results and performance analysis of the model based on prediction accuracy and computational efficiency is presented; lastly, the conclusion and future works are presented in the last section.

\section{RELATED WORK}

Over the years, different state-of-the-art prediction methods have been introduced and applied to solve travel time prediction problem. The most popular methods has been based on  Kalman Filter algorithm (KF)\cite{yangKF2005}, sequential Monte Carlo \cite{hadachi2011application}, Neural Networks (NN) and historical data analysis \cite{vivek2017}. \textit{Yang et al}, \textit{Huifeng et al} and \textit{Vanajakshi et al} explored KF methods in \cite{yangKF2005}, \cite{Huifeng2010} and \cite{vanajakshi2009} respectively. The result showed that KF methods provides sufficient solution and also have low computational cost. However, the use of KF methods are only viable in linear systems making it difficult to capture the non-linearity of travel time. In \cite{lin2013}, \textit{Lin et al} compared the performance of both NN methods and KF method, and the paper showed that NN models gives better travel time prediction accuracy than KF models. 

State of the art models based on Neural Network (NN) have always explored the use of different NN architectures and input-output combinations. For example, \textit{Duan et al} in \cite{duanLstm2016} adopted the use of Long Short-Term Memory (LSTM) architecture to solve travel time prediction problem. The prediction problem was transformed into a time series problem. LSTM ability to automatically store historical sequence was harnessed in the study to make an accurate prediction and the result reported was sufficient enough. However, the single input dimension of the proposed LSTM model means that factors that leads to travel time variability were not considered.

The use of Multilayer Perceptron (MLP) which is another type of neural network have also been explored by different researcher. In \cite{amita2015}, \textit{Amita et al} developed a MLP based travel time prediction model and the paper revealed that the model outperformed multi-linear regression model based on prediction accuracy and robustness after being tested on a GPS trajectory dataset obtained from two routes in Delhi, India.  Unlike in \cite{duanLstm2016}, the model considered three variables that causes travel time variability; dwell time, delays and distance between the stops. Although the prediction accuracy of the model was sufficient enough in this case, the transfer-ability of the proposed model can be questioned since the dataset used for training the model was collected over a period of 5 days. This means that given traffic state and infrastructural changes, the model will not be sufficient.

In addition, the use of MLP was also explored in \cite{gurmu2010dynamic} by \textit{Gurmu et al} . The developed model is based on studying both the historical and real-time arrival and departure time patterns at different stops in the selected route. The result showed that the model had $70\%$ prediction accuracy and deemed insufficient when the distance between the origin and destination is too short or too large.

Hybrid methods involving the combination of two or more models have also been proposed. \textit{Zhang et al} in \cite{Zhihaohybrid2017} developed a model with the fusion of 2-dimensional convolution neural network and LSTM. In their study, CNN model was introduced to identify the spatial features of traffic conditions from an image input with spatio-temporal characteristics while LSTM identified the correlation of the travel time series problem. The developed model performed best when compared to stand alone models like naive K-Nearest-Neighbor (KNN), historical average and instantaneous travel time. \textit{Liu et al} in \cite{Jianyinghybrid2012} took a different approach by applying the Kalman Filter to the predicted output of a model developed using Elman Neural Network. The developed hybrid model works by feeding Elman Neural Network with a historical data and the predicted output plus the observed real-time data is fed to the Kalman filter model for final prediction. Hybrid models have been proven to be more efficient when compared to other models but its high computational cost is always an issue.

\section{METHODOLOGY}

The focus of this study is on designing two different neural network models for predicting link travel times based on O-D travel time matrix generated from public bus GPS data. The methodology is divided into two: O-D travel time matrix development and Neural network model design.

\subsection{Origin-Destination Travel Time Development}

An O-D travel time matrix is a matrix which has origin stops and possible destination stops represented as rows and columns respectively. The matrix presents the distribution of travel times across links in a given trip, thus making its input important for analyzing transportation initiatives as illustrated in \cite{Shahadat2014}.
The values in the OD cell are estimated travel time between a given origin stop and a destination stop pair. This is computed using: 
\begin{equation}
TT_{o,d}^{J} = T_{Do}^{J} - T_{Ad}^{J}
\label{eqn:eqn1}
\end{equation} 

Where, $TT_{o,d}^{J}$ is the travel time in minute between any origin stop and destination stop pair. $ T_{Do}^{J}$ is the departure time at any given origin stop. $T_{Ad}^{J}$ is the arrival time at any given destination stop. 
\begin{table}[h!]
\begin{center}
\caption{Origin Destination Travel Time Matrix}
\begin{tabular}{l*{6}{c}r}
& Destination & B & C & D & E & F\\
\hline
Origin \newline \\
A                &      & $TT_{a,b}$ & $TT_{a,c}$ & $TT_{a,d}$ & $TT_{a,e}$  & $TT_{a,f}$  \\
B        &    & 0  & $TT_{b,c}$ & $TT_{b,d}$ & $TT_{b,e}$ & $TT_{b,f}$  \\
C       &    & 0 & 0 & $TT_{c,d}$ & $TT_{c,e}$  & $TT_{c,f}$  \\
D   &   & 0 & 0 & 0 & $TT_{d,e}$  & $TT_{d,f}$  \\
E  &  & 0 & 0  & 0 & 0  & $TT_{e,f}$.  \\
\end{tabular}
\label{tab:OD}
\end{center}
\end{table}

Table~\ref{tab:OD} describes the OD travel time matrix as an \textit{n x n} matrix, where \textit{n} represents the vertices of all observed stops along the trip. The weights of edges between vertices are the travel times between stops computed using equation~\ref{eqn:eqn1}. If a stop is reachable from another stop, then the travel time between the stops is placed at the \textit{ith} row and \textit{jth} column, if it is not, then 0 is placed.  

In addition, elements on the main diagonal of the matrix represents the distributed link travel time as we transverse from the origin to the final destination while the trip total travel time is calculated by finding the summation of the distributed link travel times. The resulting OD travel time matrix will provide a detailed picture of trip travel times distribution in a region and can be used by transport agencies to plan transportation needs.

\subsection{Neural Network}
Neural Network is a concept that was inspired by the operation of the brain and so far, it has been successfully applied to solve estimation and prediction problems. The interest in NN has grown over the years due to its ability to map complex non-linear relationships between variables \cite{Teresa2016}. In this section, the architectures of two neural networks used in this study are discussed.

\subsubsection{Multilayer Perceptron Network (MLP)}
MLP is a particular type of deep neural network (DNN) architecture that is often used when the relationship between input data and the expected output becomes complex. The architecture of a typical MLP neural network includes an input layer that receives external data in vector form, an hidden layer which performs input-output mapping by applying mathematical functions on the weighted inputs and an output layer that gives the result. The number of hidden layers and the number of nodes in the hidden layer that allows the network to learn more complicated features can not be predetermined before training and testing of the network.

\begin{figure}[h!]
\begin{center}
  \includegraphics[width=0.4\textwidth]{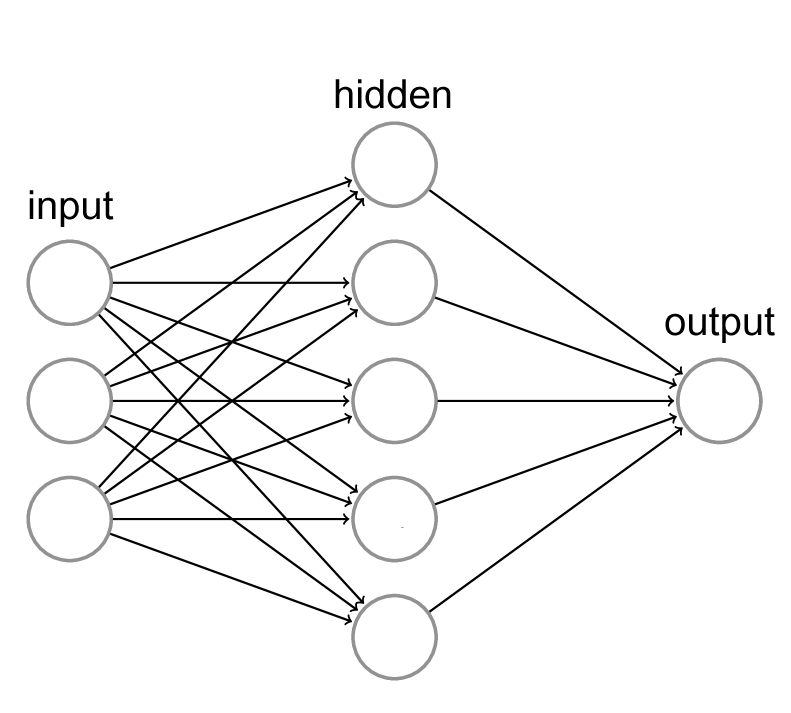}
 \caption{Multi Layer Perceptron}
 \label{fig:mlp}
\end{center}
 \end{figure}

A simple structure of MLP with 3 neurons at the input layer, 5 neurons at the hidden layer and lastly, 1 neuron at the output layer is presented in Figure~\ref{fig:mlp}. 

 Using Figure \ref{fig:mlp} as case study, the mathematical computation applied by each node in the hidden layer on weighted inputs received is: \\
\begin{equation}
Z^{[1]}={W^{[1]}}^{T}x + b
  \label{eqn:mlp}
\end{equation}

Where, $W$ is the weight of each input represented as a matrix of the form: $ W = \left[ \tiny {\begin{array}{cccc}
   w_{1,1} & w_{2,1} & w_{3,1} \\
   w_{1,2} & w_{2,2} & w_{3,2} \\
   w_{1,3} & w_{2,3} & w_{3,3} \\
   w_{1,4} & w_{2,4} & w_{3,4} \\
   w_{1,5} & w_{2,5} & w_{3,5} \\
  \end{array} } \right] $
$x$ is the input represented as a: 
$
  x= \left[ \tiny {\begin{array}{c}
   x_{1} \\
   x_{2} \\
   x_{3} \\
  \end{array} } \right]
 $ 
$b$ is the bias for each nodes in the hidden layer which is also a vector $
b = \left[ \tiny {\begin{array}{c}
   b_{1} \\
   b_{2} \\
   b_{3} \\
   b_{4} \\
   b_{5} \\
  \end{array} } \right] $
  
An activation function is applied on the result of each node to introduce non-linearity. In this case, a sigmoid activation function is used as a case study.

\begin{equation}
\sigma = \frac{1}{1 + e^{-x}}
\end{equation}
\begin{equation}
a^{[1]} = \sigma (Z^{[1]}) = \sigma ( \left[ {\begin{array}{c}
   z_{1} \\
   z_{2} \\
   z_{3} \\
   z_{4} \\
   z_{5} \\
  \end{array} } \right] )
  \label{eqn:act}
\end{equation}

The MLP network is then trained to reduce the loss function. In summary, the training algorithm is broken down into four iterative steps:

\begin{enumerate}[label=(\roman*)]
\item Read in values with random weights
\item Feed forward computation into the network
\item Get prediction $y$ and compute error $y-y_{1}$
\item Backward propagates the error into the network to update weight and bias according to an input-output deviation.
\end{enumerate}

\subsubsection{Long Short Term Memory (LSTM)}
An LSTM is a type of Recurrent Neural Network with better sophistication, longer memory and powerful transition ability than traditional recurrent neural network (RNN). The network is best known for its ability to effectively solve sequential problems such as speech recognition, image recognition and time series problems. Like MLP, the architecture of LSTM includes input, hidden and output layer, with the hidden layer nodes being fully connected. The output of a hidden layer at a particular time step $t_i$ becomes the input of the hidden layer at the next time step $t_{i+1}$.

The structure of hidden layers in LSTM is quite different from that of MLP network, with the hidden layer having memory cells that consist of gates which allows it to solve its gradient vanishing problem as it passes information from one sequence to the other. The gates are:

\begin{enumerate}[label=(\roman*)]
\item Input gate $i^{(t)}$: Takes the data sequence as an input, then applies both weight and bias on the sequenced input. 
\begin{equation}
    i^{(t)} = \sigma.(W^{ix}.x^{t} + W^{ih}.h^{(t-1)} + b^i)
\end{equation}
\item Forget gate $f^{(t)}$: This part of the memory cell decides on which prior information should be forgotten or retained in the knowledge base.
\begin{equation}
    f^{(t)} = \sigma.(W^{fx}.x^{t} + W^{fh}.h^{(t-1)} + b^f)
\end{equation}
\item Output gate $o^{(t)}$: Passes the output as an input to the input gate of the hidden layer at next time step $t_{i+1}$.
\begin{equation}
     o^{(t)} = \sigma.(W^{ox}.x^{t} + W^{oh}.h^{(t-1)} + b^f)
\end{equation}
Where all three gates are dependent on $h$ and $x$.
\end{enumerate}

\subsection{Prediction Model}

\subsubsection{Dataset}

The dataset used in this study  is an open dataset made available by Dublin City Council. It consists of GPS data collected by Dublin City Traffic Control System from different buses plying different routes in the city. It corresponds to approximately one month of data from November 6, 2012, to November 30, 2012. The dataset is popular among researchers who have worked with problems related to estimation or prediction of either bus arrival time or bus travel time. From the dataset, bus line 46A was chosen for the research described in this study. The route was chosen because it has more journey patterns when compared to other routes in the dataset. Furthermore, the route passes through Dublin city center which means it is vulnerable to peak hour traffic and travel time variation.

\subsubsection{MLP Network Prediction Model}
Based on the previously reviewed state of the art methods and the use of a data analysis tool to find the correlation between the independent and dependent variables, six input variables were considered for this model; origin stop, destination stop, distance, day, schedule and average speed. 

\subsubsection{LSTM Prediction Model}

To use an LSTM, the link travel time prediction problem was transformed into a time series problem. The focus here is to predict the total travel time of the route 46A used in this research. An LSTM neural network model was developed where at time $t_i$, the input to the LSTM network is the observed historical data at time $t_i$ while the output is the predicted travel time at $t_{i+1}$. Unlike in \cite{duanLstm2016}, in this paper, the dimension of the input at each time step isn't limited to the travel time. At each time step, the input is the travel time, the observed speed and day of the week while the dimension of the output is the dimension of the next travel time.

\section{RESULTS AND EVALUATION}
When creating NN models, there is no general selection rule for selecting number of hidden layers and neurons,  and best selection are always found through trial and error approach. However, in \cite{jeffBook}, \textit{Heaton} showed that the optimal size of hidden layer lies between the number of input and output layer. In this study, an experiment was carried out to determine the best hidden layer configuration and our experiment arrived at the structure in table \ref{tab:config}. In addition, the dataset was divided into three parts, training, validation and testing set according to the following distribution $70\%$, $15\%$ and $15\%$ respectively.

 \begin{table}[h!]
\begin{center}
\caption{NN configuration}
 \begin{tabular}{ |p{3cm} |p{1.5cm}| p{2cm} | } 
 \hline
 \textbf{Details} & \textbf{MLP Value}  & \textbf{LSTM Value} \\
 \hline
 Input Layer & 5 neuron & 3 neurons  \\ 
 \hline
 First hidden Layer  & 12 neurons & 80 neurons \\
 \hline
  Second hidden Layer  & 50 neurons & 100 neurons \\
 \hline
 Output Layer & 1 neuron & 1 neuron \\
  \hline
 Epoch & 1000 & 2000\\
   \hline
 Training Algorithm & adaGrad & adam \\
    \hline
 Batch size & - & 3 \\
 \hline
\end{tabular}
\label{tab:config}
\end{center}
\end{table}

The performance of both models were evaluated in terms of accuracy by using Root Mean Square Error (RMSE) equation \ref{eqn:estimatedTT}. 
\begin{equation}
    RMSE =  \sqrt{\frac{\sum(tt^{j}_{a,b} - TT^{j}_{a,b})^{2}}{z}}
\label{eqn:estimatedTT}
\end{equation}
Where $tt^{j}_{a,b}$ = the observed travel time from stop \textit{a} to stop \textit{b} for a journey \textit{j} in testing data. \newline
$TT^{j}_{a,b}$ = the ground truth travel time from stop \textit{a} to stop \textit{b} for a journey \textit{j} in testing data. \newline 
$z$ = the total number of test samples.

\subsection{MLP Prediction}
For MLP predictions, the OD dataset was divided into two classes, long jumps, and short jumps. Short jumps are links with short travel times within $0-35$ minutes while long jumps are trip links where the travel time from source to destination is greater than 35 minutes. 

For the last 5 days of the week, the model's performance statistics for both long and short jumps in form of RMSE was analysed. It can be seen in Table~\ref{tab:fullshort} that there is little difference between the prediction error for the two different time periods. In Table~\ref{tab:fullLong}, the prediction error for long jumps during non-peak and peak hours is presented. Again, the table shows that the model can predict long trips better with a low error margin. 
 
\begin{table}[h!]
 \caption{RMSE for short jumps}
\centering
\begin{tabular}{ |p{2cm} |p{2.5cm} |p{2.5cm} | } 
 \hline
 \textbf{Days} & \textbf{Nonpeak (minute)} & \textbf{Peak (minute)} \\
 \hline
 Monday & 1.17 & 1.18 \\ 
 \hline
 Tuesday  & 1.18 & 1.19  \\
 \hline
 Wednesday  & 1.20 & 1.16 \\
 \hline
 Thursday & 1.17 & 1.20 \\
  \hline
 Friday & 1.17 & 1.15 \\

 \hline
\end{tabular}\label{tab:fullshort}
\end{table}
\begin{table}[h!]
 \caption{RMSE for long jumps}
\centering
\begin{tabular}{ |p{2cm} |p{2.5cm} |p{2.5cm} | } 
 \hline
 \textbf{Days} & \textbf{Nonpeak (minute)} & \textbf{Peak (minute)} \\ 
 \hline
 Monday & 2.48 & 2.56  \\ 
 \hline
 Tuesday  & 2.53 & 2.58  \\
 \hline
 Wednesday  & 2.53 & 2.60 \\
 \hline
 Thursday & 2.60 & 2.61 \\
  \hline
 Friday & 2.48 & 2.52 \\

 \hline
\end{tabular}
\label{tab:fullLong}
\end{table}

\begin{figure}[h!]
\begin{center}
  \includegraphics[width=0.45\textwidth]{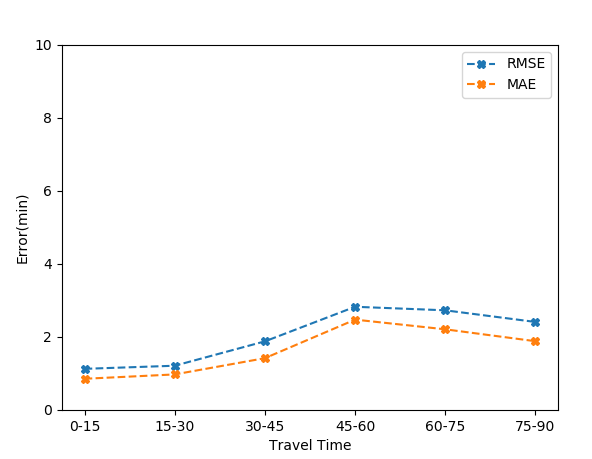}
 \caption{Error for Travel time range}
 \label{fig:meanerror}
\end{center}
 \end{figure}

In Figure~\ref{fig:meanerror}, RMSE and MAE errors for different time range is presented. From the plot, it can be seen that the model error margins increases as the travel time increases, however, it is able to predict an accurate travel time with less than $10\%$ error for anyone who is more than $30$ minutes away from the destination.

\subsection{LSTM Predictions}
\begin{figure}[h!]
\begin{center}
  \includegraphics[width=0.5\textwidth]{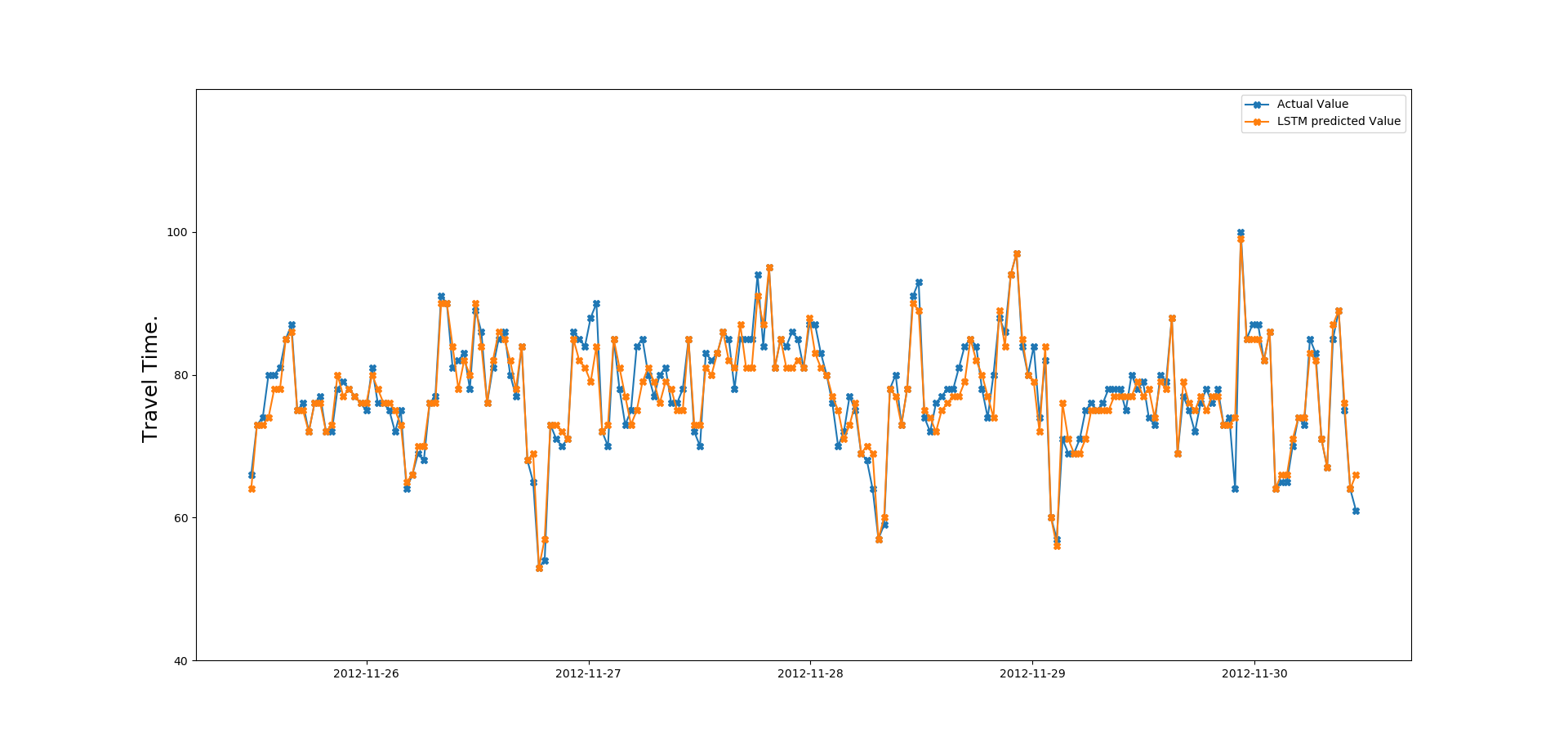}
 \caption{One time step vs Observed travel time}
 \label{fig:lstmonestep}
\end{center}
 \end{figure}

For LSTM predictions, the focus is on the full path length of the route 46A, that is,  predicting the total travel time for Stop 2039 to Stop 807. Two different time steps ($x+1$ and $x+2$) were predicted. In Figure~\ref{fig:lstmonestep}, the lstm prediction for one time step is compared to the observed travel time and it can be seen that the model is able to achieve satisfactory result. The model was able to learn both peak and non-peak periods as seen in the plot.

\begin{figure}[h]
\begin{center}
  \includegraphics[width=0.5\textwidth]{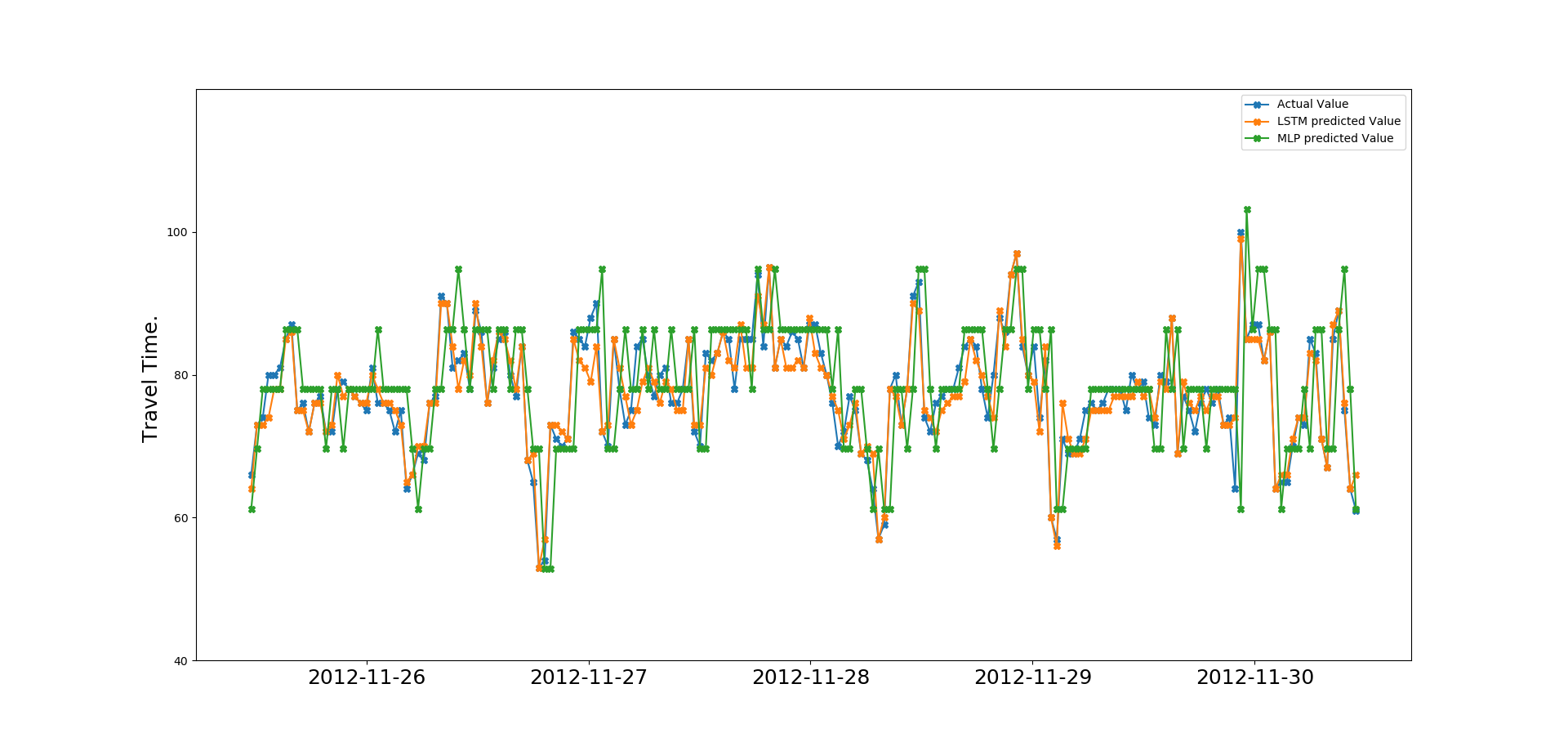}
 \caption{LSTM One time step vs MLP prediction}
 \label{fig:lstmvsmlp}
\end{center}
 \end{figure}

 In addition, the comparison between LSTM, MLP and the observed travel time is presented in Figure~\ref{fig:lstmvsmlp}. It is interesting to see that both NN models are able to learn the observed travel time sufficiently, however, the LSTM NN model's predicted travel time is a better fit to the observed travel time. This is because LSTM is best fit for sequential problem. 
 
 Lastly, in Figure~\ref{fig:lstmvsmlp} the prediction for two time step using same LSTM model is presented and it can be seen that the LSTM model in the case of two time-step is not robust against noise.  This is because the error from the previous prediction step accumulates as the next step is predicted.

 \begin{figure}[h]
\begin{center}
  \includegraphics[width=0.5\textwidth]{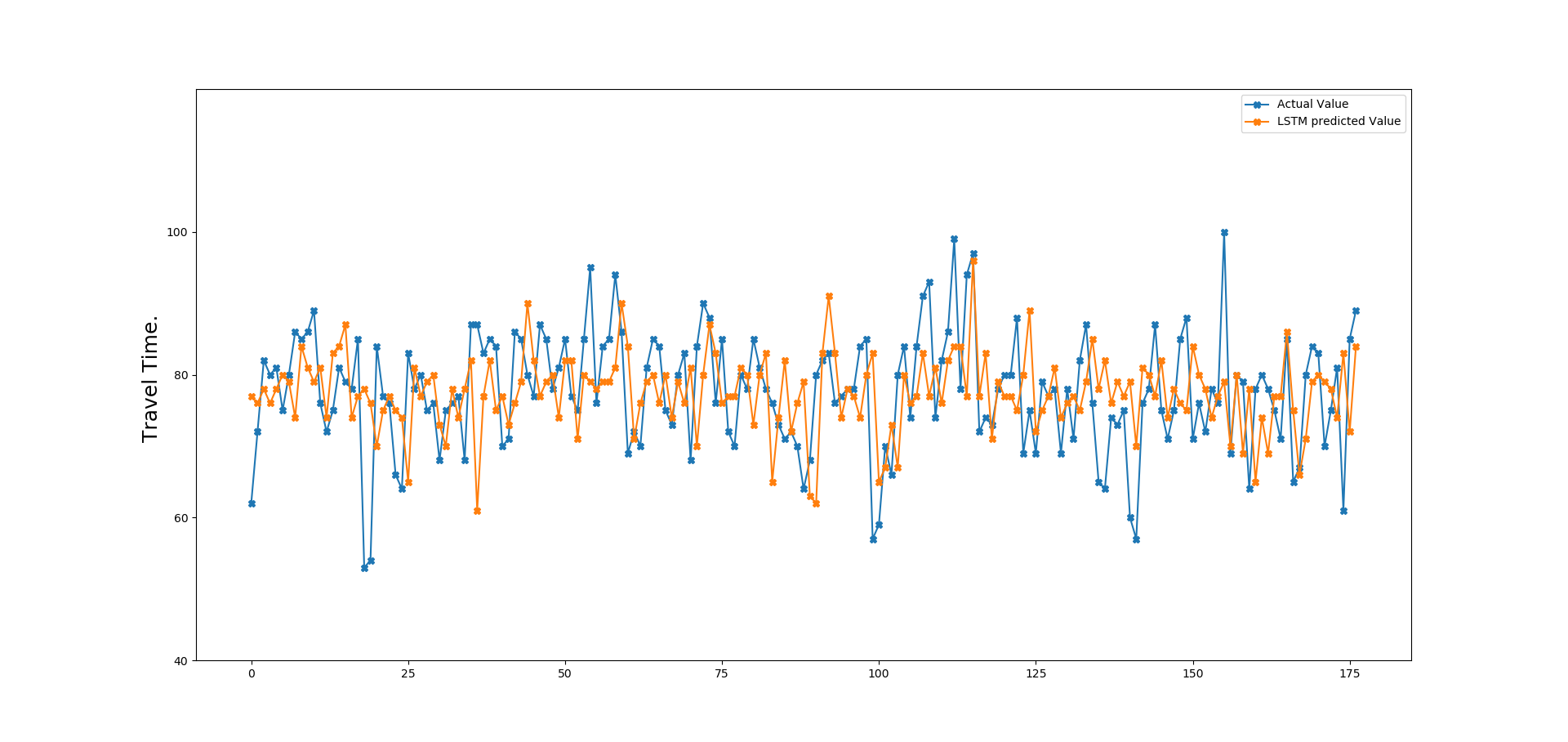}
 \caption{Two time step vs Observed travel time}
 \label{fig:lstmtwo}
\end{center}
 \end{figure}

\newpage

\section{CONCLUSIONS}

In this paper, the use of neural network (NN) and O-D travel time matrix for travel time prediction was explored. Two neural network models namely multi-layer perceptron (MLP) and long short-term model (LSTM) were built and compared based on prediction accuracy. The research result showed that both MLP and LSTM can achieve acceptable results when used in making travel time predictions. The MLP model developed produced good results for both peak and non-peak periods. Also, when the prediction accuracy of both MLP and LSTM were compared, the result showed that LSTM gives a better prediction accuracy. The research also showed that as the LSTM time steps increases, LSTM becomes susceptible to noise and this in turn reduced the prediction accuracy.  For future work, travel time data collected over a longer period of time will be used for prediction while still maintaining the neural network models. Also, noise removal techniques will be explored for the LSTM model or use convolutional neural network which is robust against noise in the data.  
\newpage
\bibliographystyle{ieeetr}
\bibliography{bibfile}

\end{document}